\documentclass{article}
\usepackage{microtype}
\usepackage{graphicx}
\usepackage{subfigure}
\usepackage{booktabs} % for professional tables
\usepackage{amsfonts}       % blackboard math symbols
\usepackage{nicefrac}       % compact symbols for 1/2, etc.
\usepackage{microtype}      % microtypography
\usepackage{amsmath}
\usepackage{empheq}
\usepackage{algorithm, algorithmic}
\usepackage[title]{appendix}
\usepackage{caption}

\usepackage{hyperref}

\usepackage[accepted]{icml2019}

\icmltitlerunning{Adversarial Example Decomposition}

\begin{document}

\twocolumn[
\icmltitle{Adversarial Example Decomposition}



\icmlsetsymbol{equal}{*}

\begin{icmlauthorlist}
\icmlauthor{Horace He}{cu}
\icmlauthor{Aaron Lou}{equal,cu}
\icmlauthor{Qingxuan Jiang}{equal,cu}
\icmlauthor{Isay Katsman}{equal,cu}
\icmlauthor{Serge Belongie}{cu}
\icmlauthor{Ser-Nam Lim}{fb}
\end{icmlauthorlist}

\icmlaffiliation{cu}{Department of Computer Science, Cornell University, Ithaca, NY, USA}
\icmlaffiliation{fb}{Facebook AI, New York, New York, USA}

\icmlcorrespondingauthor{Horace He}{hh498@cornell.edu}

\icmlkeywords{Machine Learning, ICML}

\vskip 0.3in
]



\printAffiliationsAndNotice{\icmlEqualContribution} % otherwise use the standard text.

\begin{abstract}
    Research has shown that widely used deep neural networks are vulnerable to carefully crafted adversarial perturbations. Moreover, these adversarial perturbations often transfer across models. We hypothesize that adversarial weakness is composed of three sources of bias: architecture, dataset, and random initialization. We show that one can decompose adversarial examples into an architecture-dependent component, data-dependent component, and noise-dependent component and that these components behave intuitively. For example, noise-dependent components transfer poorly to all other models, while architecture-dependent components transfer better to retrained models with the same architecture. In addition, we demonstrate that these components can be recombined to improve transferability without sacrificing efficacy on the original model.
    
    %With the recent rise of Deep Neural Networks in Computer Vision Tasks[alexNet paper?] in 2012, a similar rise in adversarial fooling has shown similar success in crippling said models. One surprising result, commonly called \textit{transferability}, demonstrates that adversarial attacks crafted to fool one model can similarly fool unrelated models. We hypothesize that the cause of adversarial weakness consists of separable factors: architecture, random initialization, and data bias. Furthermore, we show that adversarial examples can be decomposed into separate data, architecture, and noise dependent components. 
\end{abstract}

\section{Introduction}
Due to the recent successes of neural networks on a wide variety of tasks, they are now being widely applied in the real-world. However, despite their major successes, recent works have shown that in the presence of adversarially perturbed input, they fail catastrophically \cite{Szegedy2013IntriguingPO, Goodfellow2014ExplainingAH}. Moreover, \citet{Szegedy2013IntriguingPO, Goodfellow2014ExplainingAH} showed that inputs adversarially generated for one model often cause other models to misclassify images as well, a phenomenon commonly called \textit{transferability}.

Our understanding of the causes of transferability is fairly limited. \citet{Tramr2017EnsembleAT} analyzes local similarity of decision boundaries to define a local decision boundary metric that determines how transferable adversarial examples between two models are likely to be. However, many questions are still open. The recent work \citet{wu2018enhancing} hypothesized that adversarial perturbations could be decomposed into initialization-specific and data-dependent components. It is also hypothesized that the data-dependent component is primarily what contributes to transfer. However, \citet{wu2018enhancing} provides neither theoretical nor empirical evidence to justify this hypothesis.

Our work aims to examine this hypothesis in greater detail. We first augment the previous hypothesis to provide decomposition into three parts:  architecture-dependent, data-dependent, and noise-dependent components. Given this framework, our contributions are as follows:

\begin{itemize}
\item We propose a method for decomposing adversarial perturbations into noise-dependent and noise-reduced components.
\item We also present a method to further decompose the noise-reduced component into architecture-dependent and data-dependent components.
\item Extensive experiments are conducted on CIFAR-10 \cite{Krizhevsky2009LearningML} using various architectures to show the above two decompositions have the desired properties. Results from an ablation study are given to show the significance of the nontrivial choices made in our methodology. 
\end{itemize}

\begin{figure*}
  \centering
  \begin{minipage}{0.45\textwidth}
        \centering
        \includegraphics[scale=0.2]{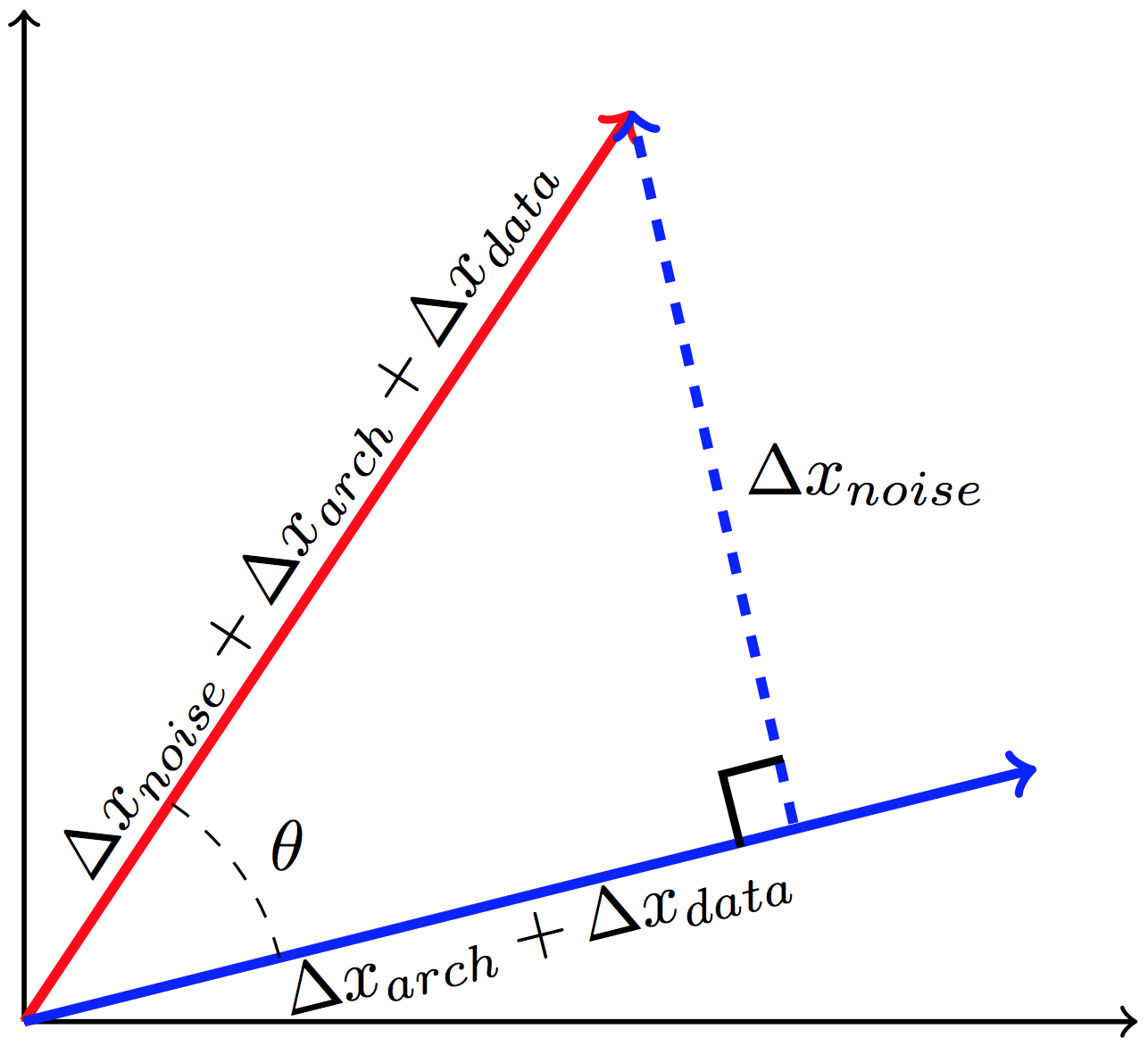}
        \caption{\textbf{Noise Vector Decomposition}. $\Delta x_{noise}$, $\Delta x_{data}$, $\Delta x_{arch}$ are as defined in Section 2.1. Note the orthogonality of $\Delta x_{noise}$ and $\Delta x_{arch} + \Delta x_{data}$; though this is only an assumption, it is justified to be reasonable experimentally in the ablation study of Section 3.3.}
    \end{minipage}\hfill
    \begin{minipage}{0.45\textwidth}
        \centering
        \includegraphics[scale=0.2]{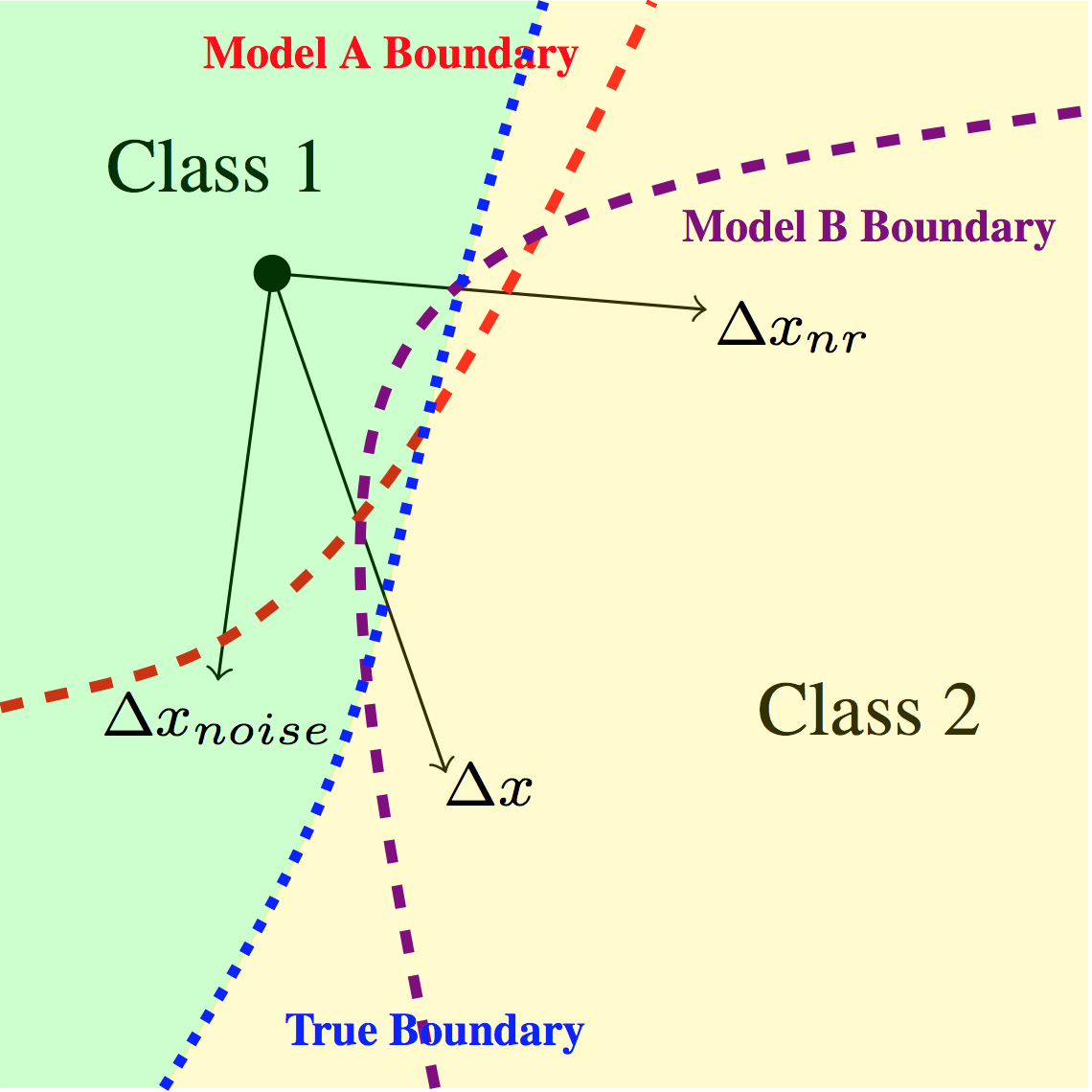}
        \caption{\textbf{Varying Decision Boundaries}. In the above figure, $\Delta x$ is the adversarial perturbation, and $\Delta x_{noise}$, $\Delta x_{nr}$ are as defined in Section 2.1.}
    \end{minipage}
  %\fbox{\rule[-.5cm]{0cm}{4cm} \rule[-.5cm]{4cm}{0cm}}
\end{figure*}
\section{Motivation and Approach}
Motivated by the reviewers' comments on \citet{wu2018enhancing}, we seek to provide further evidence that an adversarial example can be decomposed into model-dependent and data-dependent portions. First, we augment our hypothesis to claim that an adversarial perturbation can be decomposed into architecture-dependent, data-dependent, and noise-dependent components. We note that it is clear that these are the only things that could contribute in some way to the adversarial example. An intuition behind why noise-dependent components exist and would not transfer despite working on the original dataset is shown in Figure 2.

Not drawn explicitly in the figure is the architecture-dependent component. As neural networks induce biases in the decision boundary, and specific network architectures induce specific biases, we would expect that an adversarial example could exploit these biases across all models with the same architecture.

\subsection{Notation}

We denote $\mathcal{A} = \{\mathcal{A}_0,\mathcal{A}_1, \dots, \mathcal{A}_k\}$ to be the set of model architectures. Let $\mathcal{M}^i = \{\mathcal{M}_\alpha^i\}$ to be a set of fully trained models of architecture $\mathcal{A}_i$ initialized with random noise. The superscript will be omitted when architecture is clear. 

We define an attack $A(x, y, \mathcal{M}_j^i, \mathcal{L}, \Delta x_0) = \Delta x$, where $x$ is an image, $y$ its corresponding label, $\mathcal{M}_j^i$ is a neural network model as defined above, $\mathcal{L}$ is a loss function, $\Delta x_0$ the initial perturbation of $x$, and $\Delta x$ a perturbation of $x$ such that $\mathcal{L}(\mathcal{M}_j^i(x + \Delta x), y)$ is maximal. 

For fixed architecture $\mathcal{A}_i$, model $\mathcal{M}_{j}^{i}$, and attack $A$, we denote $\Delta x_{noise}, \Delta x_{arch}, \Delta x_{data}$ to be the three components of $\Delta x$ introduced in previous sections. Let $\Delta x_{noise \ reduced} = \Delta x_{arch} + \Delta x_{data}$; we will use the short hand $\Delta x_{nr}$. 

% We define intra-architecture transferability to refer to transferability among models sharing the same architectures, and inter-architecture transferability to refer to transferability among models with different architectures.

Let $P_{x_{1}}(x_{2})$ denotes the projection of vector $x_{2}$ onto vector $x_{1}$. Let $\widehat{x}=\frac{x}{||x||}$ be the unit vector with same direction as $x$. 

%We denote a fixed model architecture with $\mathcal{A}_i$, and a trained model that is an output of a single training run of $\mathcal{A}_i$ as $\mathcal{M}^i_j$. The superscript will be omitted when the architecture is clear from context. 
%We define an attack method $A$ to take in two inputs: initial perturbation $\Delta x_{0}$ and loss function $L_{\mathcal{M}}(x+\Delta x, y)$ (depending on model $\mathcal{M}$ and data point $(x,y)$). Attack $A$ starts with $\Delta x=\Delta x_{0}$ and finds the optimal $\Delta x$ that minimizes this loss function under certain constraints. 

%Let $x^* = x + \Delta x$ be an adversarial example for model $\mathcal{M}_1$, where $\Delta x$ is the output of an attack method $A$ with initial perturbation $\Delta x_{0}=0$ and loss function $L_{\mathcal{M}_{1}}$ with parameter $M_{1}$. We consider $\Delta x$ to be composed of 3 additive components $\Delta x_{noise}$, $\Delta x_{arch}$, and $\Delta x_{data}$ (ie: $\Delta x = \Delta x_{noise} + \Delta x_{arch} + \Delta x_{data}$. A perturbation consisting of $\Delta x_{arch} + \Delta x_{data}$ is denoted $\Delta x_{noise\ reduced}$ or $\Delta x_{nr}$. 
%We use the term inter-architecture transferability to refer to transferability among two models with the same architecture, and the term intra-architecture transferability to refer to transferability among two models with different architectures. 
%$P_{x_{1}}(x_{2})$ denotes the projection of vector $x_{2}$ onto vector $x_{1}$. 

\subsection{$\Delta x_{noise}$ and $\Delta x_{nr}$ Decomposition}
\label{noisemethod}
\textbf{Description: } We fix our architecture $\mathcal{A}_0$ and have $\{\mathcal{M}_1, \dots, \mathcal{M}_n\}$ as our set of trained models. Set $\mathcal{L}$ to be the cross-entropy loss and let $\Delta x = A(x, y, \mathcal{M}_1, \mathcal{L}, \mathbf{0})$ be the generated adversarial perturbation for $\mathcal{M}_1$. 

\textbf{Proposition: } $\Delta x$ can be decomposed into $\Delta x_{noise} + \Delta x_{nr}$ such that the attack $\Delta x_{noise}$ is effective on $\mathcal{M}_1$ but transfers poorly to $\mathcal{M}_2, \dots, \mathcal{M}_n$, while $\Delta x_{nr}$ transfers well on all models.

The equations for computing $\Delta x_{nr}$ and $\Delta x_{noise}$ are given in Equation \ref{eq:xnoisexnrdecomp} (see Appendix C for justification). The technique is illustrated in Figure 1. 
\begin{equation} \label{eq:xnoisexnrdecomp}
\begin{gathered}
\Delta x_{nr} \approx A(x, y, \frac{1}{n - 1} \displaystyle{\sum_{j = 2}^n}\mathcal{M}_j, \mathcal{L}, \mathbf{0}) \\
\Delta x_{noise} \approx \Delta x-P_{\Delta x_{nr}}(\Delta x)
\end{gathered}
\end{equation}

\subsection{$\Delta x_{arch}$ and $\Delta x_{data}$ Decomposition}
\label{modelmethod}
\textbf{Description: } We reuse notation from the above section, except that we now consider a set of different architectures $\mathcal{A} = \{\mathcal{A}_0, \mathcal{A}_1, \dots \mathcal{A}_k\}$

\textbf{Proposition:} $\Delta x_{nr}$ can be composed into $\Delta x_{arch} + \Delta x_{data}$ such that the attack $\Delta x_{arch}$ is effective on $\mathcal{A}_0$ but transfers poorly to $\mathcal{A}_{1}, \cdots, \mathcal{A}_{k}$, while $\Delta x_{data}$ transfers well on all models.

The equations for computing $\Delta x_{arch}$ and $\Delta x_{data}$ are given in Equation 2, in which we set $\Delta x_{nr}$ to be the noise reduced perturbation generated on $\mathcal{A}_0$ (see Appendix C for justification). We approximate the expectation for $\Delta x_{data}$ by averaging across architectures.
\begin{equation} \label{eq:xarchxdatadecomp}
\begin{gathered}
\Delta x_{data} \approx \\
\mathbb{E}_{\mathcal{A}_i \in \mathcal{A}} \left[A(x, y, \frac{1}{n - 1} \sum_{j = 2}^n \mathcal{M}_j^i, \mathcal{L} , \Delta x_{nr})\right] \\
\Delta x_{arch} \approx \Delta x_{nr} - P_{\Delta x_{data}} (\Delta x_{nr})
\end{gathered}
\end{equation}

\begin{table*}[!htb]
  \begin{minipage}{.55\linewidth}
  \caption{$\Delta x_{noise}$ decomposition (ResNet18)}
  \label{noisedependentres}
  \centering
  \begin{tabular}{lccc}
    \toprule
    $\Delta$ & $\mathcal{M}_{orig}$ & $\{\mathcal{M}_{avg}\}$ & $\{\mathcal{M}_{test}\}$ \\
    \midrule
    $\Delta x$ & 68.3\% & 45.6\% & 46.7\%    \\
    $\Delta x_{nr}$  & 63.7\% & 61.9\% & 59.5\%   \\
    $\Delta x_{noise}$ & 60.2\% & 19.8 \% & 20.3\% \\
    \bottomrule
  \end{tabular}
  \end{minipage}
  \begin{minipage}{.38\linewidth}
  \caption{$\Delta x_{noise}$ decomposition (DenseNet121)}
  \label{noisedependentdense}
  \begin{tabular}{lccc}
    \toprule
    $\Delta$ & $\mathcal{M}_{orig}$ & $\{\mathcal{M}_{avg}\}$ & $\{\mathcal{M}_{test}\}$ \\
    \midrule
    $\Delta x$ & 70.0 \% & 47.1\% & 49.8\% \\
    $\Delta x_{nr}$  & 64.3\% & 65.3\% & 66.6\%  \\
    $\Delta x_{noise}$ & 64.9\% & 27.1\% & 29.6\% \\
    \bottomrule
  \end{tabular}
  \end{minipage}
  \vspace{0.3cm}
    \caption*{All numbers reported are fooling ratios. Observe that $\Delta x_{noise}$ exhibits exceptionally low transferability.}
\end{table*}

\section{Results}
We empirically verify the approaches given in the motivation above and show that the isolated noise and architecture-dependent perturbations show the desired properties. Unless stated otherwise, all perturbations are generated on CIFAR-10 \cite{Krizhevsky2009LearningML} (original images rescaled to $[-1,1]$) using iFGSM \cite{Kurakin2016AdversarialEI} with 10 iterations, distance metric $L_{\infty}$, and $\epsilon=0.03$. All experiments are run on the first 2000 CIFAR-10 test images. In addition, all models are trained for only 10 epochs due to computational constraints. All percentages reported are fooling ratios \cite{MoosaviDezfooli2017UniversalAP}. For results with other settings, check Appendix A.

\subsection{$\Delta x_{noise}$ and $\Delta x_{nr}$ Decomposition}
We start off with a set of 10 retrained ResNet18 \cite{He2016DeepRL} models $\{\mathcal{M}_i\}$. We attack the first ResNet18 model $\mathcal{M}_1$($=\mathcal{M}_{orig}$) to get a perturbation $\Delta x$ for a given $x$. We then follow the process outlined in Equation \ref{eq:xnoisexnrdecomp} to obtain $\Delta x_{nr}$ from the other 9 retrained ResNet18 models $\{\mathcal{M}_{i > 1}\}$($=\{\mathcal{M}_{avg}\}$). We then test on an untouched set of 5 retrained ResNet18 models $\{\mathcal{M}_{test}\}$. We also do the same process for DenseNet121 \cite{Huang2017DenselyCC} instead of ResNet18 and report their respective results in Tables 1 and 2.

We note that $\Delta x_{noise}$ achieves a far lower transfer rate than either $\Delta x_{nr}$ or $\Delta x$ while still maintaining relatively high error rate on the original model, providing evidence for the success of this decomposition. To the best of our knowledge, this is the first methodology that is able to construct adversarial examples with especially low transferability. Although this is of low practical use, this is theoretically interesting. We note that although we attempt to generate $\Delta x_{nr}$ by multi-fooling across 9 retrained models, reducing noise in high dimensions is difficult, so we are unable to achieve a perfect decomposition of $\Delta x_{noise}$. Ablation studies in Appendix B suggest that we may be able to achieve a better decomposition with a larger set of retrained models.

\subsubsection{Recombining components}
As the components $\Delta \widehat{x}_{noise}$ and $\Delta \widehat{x}_{nr}$ are linearly independent unit vectors, and by definition, $\Delta x$ is in the span of these vectors, we can find unique scalars $a$ and $b$ such that $a\cdot\Delta \widehat{x}_{noise} + b\cdot\Delta\widehat{x}_{nr} = \Delta x$. Experimentally, we find that under our setting, $a\approx 1.319$ and $b\approx 0.386$. We note that for our original perturbation, this is perhaps an undue amount of focus paid to the noise-specific perturbation. 
We can now try setting $a$ and $b$ to different ratios, which correspond to how much we wish to emphasize attacking the original model vs. transferability. As we are now able to set an arbitrarily high $a$ and $b$, allowing us to saturate the epsilon constraints, we sign maximize (ie: $sign(x)\cdot\epsilon$), as motivated in \citet{Goodfellow2014ExplainingAH}) to level the playing field. The results in Table 3 show the results of performing these experiments on ResNet18. We find that we are able to generate perturbations that perform equivalently with $\Delta x$ on $\mathcal{M}_{orig}$, while performing substantially better when transferring to $\{\mathcal{M}_{avg}\}$ and $\mathcal{M}_{test}$.
\begin{table}[!htb]
  \caption{Linear Combinations of $\Delta \widehat{x}_{noise}$ and $\Delta \widehat{x}_{nr}$}
  \label{linearcombination}
  \centering
  \begin{tabular}{ccccc}
    \toprule
    $b:a$ &  $\mathcal{M}_{orig}$ & $\{\mathcal{M}_{avg}\}$ & $\mathcal{M}_{test}$ \\
    \midrule
    $\Delta x_{nr}$ & 65.8\% & 63.6\% & 65.1\%    \\
    2:1 & 68.5\% & 63.7\% & 65.2\%    \\
    % 1.75:1 & 68.8\% & 63.5\% & 64.7\%    \\
    1.5:1 & 69.4\% & 61.2\% & 62.8\%    \\
    % 1.25:1 & 69.4\% & 58.5\% & 58.4\%    \\
    1:1 & 69.8\% & 56.0\% & 56.4\%    \\
    % 1:1.25 & 69.9\% & 55.1\% & 55.1\%    \\
    1:2 & 70.0\% & 53.1\% & 53.5\%    \\
    % 1:4 & 70.0\% & 50.4\% & 50.6\%    \\
    $\Delta x$ & 69.8\% & 51.0\% & 51.0\%    \\
    \bottomrule
  \end{tabular}
    \caption*{All numbers reported are fooling ratios. Observe that as you increase the ratio $b:a$ we obtain better transferability with lowered effectiveness on $\mathcal{M}_{orig}$. Also note that we are able to construct perturbations that are strictly superior to either $\Delta x$ or $\Delta x_{nr}$. }
\end{table}

\begin{table*}[!htb]
  \caption{$\Delta x_{arch}$ decomposition}
  \label{tab:archdecomp}
  \centering
  \begin{tabular}{clcccc}
    \toprule
    source & $\Delta$ & \{$\mathcal{M}_{source}\}$ & $\{\mathcal{M}_{other}\}$ & $\{\mathcal{M'}_{source}\}$ & $\{\mathcal{M'}_{other}\}$ \\
    \midrule
    & $\Delta x_{nr}$ & 60.9\% & 50.7\% & 59.4\% & 50.7\%   \\
    ResNet18 & $\Delta x_{data}$  & 54.6\% & 61.4\% & 54.8\%  & 60.8\% \\
    & $\Delta x_{arch}$ & 52.4\% & 26.7\% & 36.9\% & 30.3\% \\
    \midrule
    & $\Delta x_{nr}$ & 62.8\% & 46.2\% & 62.9\% & 47.1\%   \\
    DenseNet121 & $\Delta x_{data}$  & 58.4\% & 58.3\% & 57.2\%  & 55.7\% \\
    & $\Delta x_{arch}$ & 54.1\% & 24.4\% & 43.1\% & 26.0\% \\
    \midrule
    & $\Delta x_{nr}$ & 65.3\% & 41.9\% & 65.7\% & 41.9\%   \\
    GoogLeNet & $\Delta x_{data}$  & 59.5\% & 59.2\% & 59.5\%  & 58.3\% \\
    & $\Delta x_{arch}$ & 57.9\% & 22.8\% & 44.8\% & 26.2\% \\
    \midrule
    & $\Delta x_{nr}$ & 53.8\% & 48.4\% & 53.2\% & 49.0\% \\
    SENet18 & $\Delta x_{data}$  & 55.7\% & 64.5\% & 54.8\%  & 63.8\% \\
    & $\Delta x_{arch}$ & 47.1\% & 28.1\% & 38.6\% & 29.8\% \\
    \bottomrule
  \end{tabular}
  \caption*{All numbers reported are fooling ratios. Note that for all architectures, the adversarial decomposition holds. Namely, $\Delta x_{arch}$ is more transferable to its specific architecture than to others, whereas $\Delta x_{data}$ is equally transferable across architectures.}
\end{table*}

\subsection{$\Delta x_{arch}$ and $\Delta x_{data}$ Decomposition}
% \textcolor{red}{Fix redundancy with above stuff.}
To evaluate decomposition into architecture and data-specific components, we consider the four architectures ResNet18 \cite{He2016DeepRL}, GoogLeNet \cite{Szegedy2015GoingDW}, DenseNet121 \cite{Huang2017DenselyCC}, and SENet18 \cite{Hu2017SqueezeandExcitationN}. Results are given in Table \ref{tab:archdecomp}. In each experiment we first fix a source architecture $\mathcal{A}_i$ and generate $\Delta x_{nr}$ by attacking 4 retrained copies of $\mathcal{A}_i$, denoted as $\{\mathcal{M}_{source}\}$. We then generate $\Delta x_{data}$  by attacking four copies of each $\mathcal{A}_{j\neq i}$ for twelve models total. We then test on another 4 retrained copies of $\mathcal{A}_i$ called $\{\mathcal{M}'_{source}\}$ as well as $\{\mathcal{M}'_{other}\}$, consisting of four copies of each of the other three architectures $\{\mathcal{A}_{j\neq i}\}$.
We see that for all four models, $\Delta x_{arch}$ obtains significantly higher error rate on $\{\mathcal{M}'_{source}\}$ than on $\{\mathcal{M}'_{other}\}$. In addition, the relative error between $\{\mathcal{M}'_{source}\}$ and $\{\mathcal{M}'_{other}\}$ for $\Delta x_{arch}$ are close to the relative error between $\{\mathcal{M}_{source}\}$ and $\{\mathcal{M}_{other}\}$ for $\Delta x_{nr}$ when averaged across models, supporting the success of our decomposition.
%However, as we can see from the large difference in error rate for $\Delta x_{arch}$ between $\{\mathcal{M}_{source}\}$ and $\{\mathcal{M}'_{other}\}$, $\Delta x_{arch}$ still must have a significant $\Delta x_{noise}$ component. We hypothesize that this is due to $||\{\mathcal{M}_{other}\}||$ being relatively low. Although we'd like to obtain $\Delta x_{data}$ from attacking this set of models, the limited size of that set (both in terms of number of models as well as architectures), means that our decomposition retains plenty of noise that is unable to be isolated.

\subsection{Ablation}
\textbf{Orthogonality}
We assume that $\Delta x_{noise}$, $\Delta x_{arch}$, and $\Delta x_{data}$ terms are orthogonal. We note that if these vectors had no relation to each other, then due to the properties of high dimensional space, they are approximately orthogonal with very high probability. 

We vary orthogonality by modifying the method in Section \ref{noisemethod} to generate $\Delta x_{noise}$ with $\Delta x - \alpha P_{\Delta x_{nr}}(\Delta x)$. When $\alpha = 1$, we recover the original algorithm, and when $\alpha=0$, $\Delta x_{noise} = \Delta x$.
Experimentally varying the orthogonality of $\Delta x_{noise}$ and $\Delta x_{nr}$ produces the results in Table \ref{tab:alpha-table}; note that we achieve the greatest difference in efficacy between the original model and transferred models when they are near-orthogonal, suggesting that the assumption we made is reasonable.

However, it is not true that orthogonal components achieve the best isolation (given by the fact that the peak difference seems to be at $\alpha = 1.2$). This suggests that our current method of decomposition may simply be an approximation for the true components, and that a more nuanced method may be necessary for better isolation.

\begin{table}[!htb]
  \caption{Varying $\alpha$}
  \label{tab:alpha-table}
  \centering
  \begin{tabular}{ccccc}
    \toprule
    $\alpha$ & $\mathcal{M}_{orig}$ & $\{\mathcal{M}_{avg}\}$ & Difference \\
    $\Delta x$ & 68.3\% & 45.6\% & 22.7\\
    $\Delta x_{nr}$ & 63.7\% & 61.9\% & 1.8\\
    0.1 & 66.7\% & 42.2\% & 24.5\\
    0.5 & 63.4\% & 29.1\% & 34.3\\
    0.8 & 59.7\% & 21.4\% & 38.3\\
    1.0 & 52.7\% & 16.6\% & 36.1\\
    1.2 & 51.9\% & 11.3\% & 40.6\\
    1.5 & 42.9\% & 9.4\% & 33.5\\
    2.0 & 33.5\% & 7.2\% & 26.3\\
    \bottomrule
  \end{tabular}
  \caption*{All percentages reported are fooling ratios. Note that the $\alpha=1.2$ setting is what produces maximal difference, which is slightly different from the assumed orthogonality ($\alpha=1.0$).}
\end{table}

\textbf{Number of Models}
We find that the higher the number of models we use to approximate $\Delta x_{nr}$, the more successfully we are able to isolate $\Delta x_{noise}$. Check Appendix B for full results.

\section{Conclusion}
    We demonstrate that it is possible to decompose adversarial perturbations into noise-dependent and data-dependent components, a hypothesis reviewers thought was interesting but unsupported in \cite{wu2018enhancing}. We go even further by decomposing an adversarial perturbation into model related, data related, and noise related perturbations. A major contribution here is a new method of analyzing adversarial examples; this creates many potential future directions for research. One interesting direction would be extending these decompositions to universal perturbations \cite{MoosaviDezfooli2017UniversalAP, Poursaeed2017GenerativeAP} and thus removing the dependence on individual data points. Another avenue to explore is analyzing various attacks and defenses and how they interplay with these various components.
% We think extensions of such interpretable decompositions to the setting of universal perturbations \cite{MoosaviDezfooli2017UniversalAP, Poursaeed2017GenerativeAP} would make for interesting future work.

\newpage
\bibliography{example_paper}

\clearpage
\newpage

\renewcommand\appendixpagename{Appendix}
\appendixpage
\section*{A. Different attack settings}
To show that our decomposition is effective across a variety of attack settings, we perform the experiment of Section 3.1 with three different iFGSM settings corresponding to $\epsilon=0.01,0.03,0.06$. Results are shown in Table \ref{tab:varying-eps}.

\begin{table}[!htb]
  \caption{Varying $\epsilon$}
  \label{tab:varying-eps}
  \centering
  \begin{tabular}{clcccc}
    \toprule
    $\epsilon$ & $\Delta$ & \{$\mathcal{M}_{orig}\}$ & $\{\mathcal{M}_{avg}\}$ & $\{\mathcal{M}_{test}\}$ \\
    \midrule
    & $\Delta x$ & 39.0\% & 16.4\% & 14.4\% \\
    .01 & $\Delta x_{nr}$  & 25.1\% & 28.4\% & 22.2\%  \\
    & $\Delta x_{noise}$ & 26.6\% & 06.2\% & 05.3\% \\
    \midrule
    & $\Delta x$ & 68.3\% & 45.6\% & 46.7\% \\
    .03 & $\Delta x_{nr}$ & 63.7\% & 61.9\% & 59.5\% \\
    & $\Delta x_{noise}$ & 60.2\% & 19.8\% & 20.3\% \\
    \midrule
    & $\Delta x$ & 81.2\% & 69.7\% & 73.6\%  \\
    .06 & $\Delta x_{nr}$  & 81.1\% & 80.5\% & 85.8\% \\
    & $\Delta x_{noise}$ & 77.7\% & 39.4\% & 40.0\% \\
    \bottomrule
  \end{tabular}
\end{table}
\section*{B. Varying number of models/iterations}
We investigate the effectiveness of the Section 3.1 decomposition as we vary hyper-parameters. Results for increasing iFGSM iterations in Table \ref{tab:ifgsm-iters} and results for increasing the results for increasing the number of models are give in Table \ref{tab:vary-models}.

\begin{table}[!htb]
  \caption{Varying number of iterations used for iFGSM}
  \label{tab:ifgsm-iters}
  \centering
  \begin{tabular}{clcccc}
    \toprule
    \# of iters & $\Delta$ & \{$\mathcal{M}_{orig}\}$ & $\{\mathcal{M}_{avg}\}$ \\
    \midrule
    & $\Delta x$ & 65.2\% & 43.4\% \\
    5 & $\Delta x_{nr}$  & 58.8\%& 58.8\%  \\
    & $\Delta x_{noise}$ & 55.5\% & 20.6\% \\
    \midrule
    & $\Delta x$ & 68.3\% & 46.7\%  \\
    10 & $\Delta x_{nr}$  & 63.7\% & 61.9\%  \\
    & $\Delta x_{noise}$ & 60.2\% & 19.8\%  \\
    \midrule
    & $\Delta x$ & 72.9\% & 48.6\%   \\
    100 & $\Delta x_{nr}$  & 67.3\% & 65.2\%  \\
    & $\Delta x_{noise}$ & 60.3\% & 18.7\% \\
    \bottomrule
  \end{tabular}
\end{table}
\begin{table}[!htb]
  \caption{Varying number of models used to approximate $\Delta x_{nr}$}
  \label{tab:vary-models}
  \centering
  \begin{tabular}{clcccc}
    \toprule
    \# of models & $\Delta$ & \{$\mathcal{M}_{orig}\}$ & $\{\mathcal{M}_{avg}\}$ & $\{\mathcal{M}_{test}\}$ \\
    \midrule
    & $\Delta x$ & 69.4\% & 46.6\% & 45.6\% \\
    3 & $\Delta x_{nr}$  & 57.6\% & 62.1\% & 51.9\%  \\
    & $\Delta x_{noise}$ & 60.1\% & 24.9\% & 29.2\% \\
    \midrule
    & $\Delta x$ & 68.4\% & 47.0\% & 44.8\% \\
    5 & $\Delta x_{nr}$  & 60.1\% & 62.0\% & 55.2\% \\
    & $\Delta x_{noise}$ & 57.5\% & 22.4\% & 24.6\% \\
    \midrule
    & $\Delta x$ & 68.3\% & 45.6\% & 46.7\%  \\
    10 & $\Delta x_{nr}$  & 63.7\% & 61.9\% & 59.5\% \\
    & $\Delta x_{noise}$ & 60.2\% & 19.8\% & 20.3\% \\
    \bottomrule
  \end{tabular}
\end{table}

\section*{C. Justification of Equations}

\textbf{Justification of Equations in 3.1}

\noindent Recall that the equations are given by

$$\begin{gathered}
\Delta x_{nr} \approx A(x, y, \frac{1}{n - 1} \displaystyle{\sum_{j = 2}^n}\mathcal{M}_j, \mathcal{L}, 0) \\
\Delta x_{noise} \approx \Delta x-P_{\Delta x_{nr}}(\Delta x)
\end{gathered}$$

We assume that the expected value of our noise term $\Delta x_{noise}$ is $0$ over all random noise. This is motivated because the random noise $i$ at initialization is a Gaussian distribution centered at $0$, and it is reasonable to assume that the model distribution and the noise distribution follows a similar pattern.

Letting $\Delta x^j = A(x, y, \mathcal{M}_j, \mathcal{L}, 0)$ over all random initialization $i$, we claim that $\mathbb{E}_j[\Delta x^j] = \Delta x_{arch} + \Delta x_{data}$. Since $\Delta x_{arch}$ and $\Delta x_{data}$ are noise independent, which means that

$$\Delta x^j = \Delta x_{noise}^j + \Delta x_{arch} + \Delta x_{data}$$

where $\Delta x_{noise}^j$ is the noise component corresponding with the noise of model $\mathcal{M}_j$. Therefore, it follows that 

$$\begin{gathered}\mathbb{E}_j[\Delta x^j] = \Delta x_{arch} + \Delta x_{data} + \mathbb{E}_j[\Delta x_{noise}^j] \\ 
= \Delta x_{arch} + \Delta x_{data} = \Delta x_{nr}\end{gathered}$$

By the law of large numbers, it follows that $\displaystyle{\lim_{n \to \infty} \frac{1}{n} \sum_{j = 1}^n} \Delta x^j = \Delta x_{nr}$. Therefore, we note that, for sufficiently large $n$, it follows that

$$\frac{1}{n} \sum_{j = 1}^n \Delta x^j \approx \Delta x_{nr}$$

We see that, since the cross entropy loss $\mathcal{L}$ is additive and the attack $A$ that we examine are first order differentiation methods, we have

$$\begin{gathered}\mathcal{L} (\frac{1}{n - 1} \sum_{j = 2}^n \mathcal{M}_j (x), y) = \frac{1}{n - 1} \sum_{j = 2}^n \mathcal{L} (\mathcal{M}_j (x), y) \\ \implies A(x, y,  \frac{1}{n - 1} \sum_{j = 2}^n \mathcal{M}_j, \mathcal{L}, 0) \\ = \frac{1}{n - 1} \sum_{j = 2}^n A(x, y, \mathcal{M}_j, \mathcal{L}, 0) \approx \Delta x_{nr}\end{gathered}$$

To prove the other claim, we have already shown through empirical results and an intuition that $\Delta x_{noise}$ and $\Delta x_{nr}$ are linearly independent that $\Delta x_{noise}$ and $\Delta x_{nr}$ are very close to orthogonal and compose $\Delta x$. Therefore, it follows that we can take the use the projection of $\Delta x_{nr}$ implies that

$$\begin{gathered}\Delta x_{noise} + P_{\Delta x_{nr}} (\Delta x) \approx \Delta x \\ \implies \Delta x_{noise} \approx \Delta x - P_{\Delta x_{nr}}(\Delta x)\end{gathered}$$

up to a scaling constant. \\

\textbf{Justification of Equations in 3.2}

Recall that the equations are, given $\Delta_{nr}$ generated on $\mathcal{A}_0$,

$$\begin{gathered}\Delta x_{data} \approx \mathbb{E}_{\mathcal{A}_i \in \mathcal{A}} \left[A(x, y, \frac{1}{n - 1} \sum_{j = 2}^n \mathcal{M}_j^i, \mathcal{L} , \Delta x_{nr})\right] \\ 
\Delta x_{arch} \approx \Delta x_{nr} - P_{\Delta x_{data}} (\Delta x_{nr})\end{gathered}$$

We make two core assumptions: 

\begin{itemize}
    \item The value of $\Delta \mathbb{E}_\mathcal{A} [x_{arch}] = 0$. This is a reasonable assumption since our generated architectures $\mathcal{A}$ should produce roughly symmetric error vectors $x_{arch}$. 
    
    \item $A(x, y, \frac{1}{n - 1} \sum_{j = 2}^n\mathcal{M}, \mathcal{L}, \Delta x')$ is equivalent $A(x, y, \mathcal{M}, \mathcal{L}, 0)$ in the sense that the former produces a noised reduce gradient closer to $\Delta x'$. This is reasonable because the space of there are many adversarial perturbations (different directions) and changing our start location won't cripple our search space. Furthermore, we use this to generate a $\Delta_{nr}$ close to $\Delta x'$. 
\end{itemize}

We claim that $\mathbb{E}_\mathcal{A} [\Delta x_{nr}] = \Delta x_{data}$ where we take $\Delta x_{nr}$ over architecture $\mathcal{A}$. To see this, we note that 

$$\mathbb{E}_\mathcal{A} [\Delta x_{nr}] = \mathbb{E}_\mathcal{A} [\Delta x_{arch}] + \mathbb{E}_\mathcal{A} [\Delta x_{data}] = \mathbb{E}_\mathcal{A}[\Delta x_{data}]$$

and so again we can approximate it with $\displaystyle{\lim_{n \to \infty}} \frac{1}{n} \sum_{i = 1}^n \Delta x_{nr}^i = \Delta x_{data}$ where $\Delta x_{nr}^i$ is the $\Delta x_{nr}$ component generated for model $\mathcal{A}_i$. For sufficiently large $n$, it follows that

$$\frac{1}{n} \sum_{i = 1}^n \Delta x_{nr}^i = \Delta x_{data}$$

Therefore we have

$$\begin{gathered}\Delta x_{data} = \mathbb{E}_{\mathcal{A}_i \in \mathcal{A}}[\Delta x_{nr}^i] = \mathbb{E}_{\mathcal{A}_i \in \mathcal{A}} \mathbb{E}_{j} [(x, y, \mathcal{M}_j^i, \mathcal{L}, 0)] \\ \approx \mathbb{E}_{\mathcal{A}_i \in \mathcal{A}} \left[A(x, y, \frac{1}{n - 1} \sum_{j = 2}^n \mathcal{M}_j^i, \mathcal{L} ,0)\right]\end{gathered}$$

and by our assumption this is roughly equivalent to

$$\mathbb{E}_{\mathcal{A}_i \in \mathcal{A}} \left[A(x, y, \frac{1}{n - 1} \sum_{j = 2}^n \mathcal{M}_j^i, \mathcal{L} , \Delta x_{nr})\right]$$

as desired. To prove the other claim, we use an analogous argument to the one above as we have shown that $\Delta x_{arch}$ and $\Delta x_{data}$ are orthogonal and applying the same projection technique yields

$$\begin{gathered}\Delta x_{data} + P_{\Delta x_{data}}(\Delta x_{nr}) \approx \Delta x_{nr} \\ \implies \Delta x_{arch} \approx \Delta x - P_{\Delta x_{data}}(\Delta x_{nr})\end{gathered}$$

up to a scaling constant.

\end{document}